\documentclass[letterpaper,twocolumn,fleqn]{article} 

\usepackage{ist}
\usepackage{times}
\usepackage{soul}
\usepackage{url}
\usepackage[utf8]{inputenc}
\usepackage[small]{caption}
\usepackage{graphicx}
\usepackage{amsmath}
\usepackage{amsthm}
\usepackage{booktabs}
\usepackage{algorithm}
\usepackage{algorithmic}
\usepackage{epsfig}
\usepackage{subcaption}
\usepackage{amssymb}

\title{An End-to-End Food Image Analysis System}

\author{Jiangpeng He\textsuperscript{1}, Runyu Mao\textsuperscript{1}, Zeman Shao\textsuperscript{1}, Janine L. Wright\textsuperscript{2}, Deborah A. Kerr\textsuperscript{2}, Carol J. Boushey\textsuperscript{3}, Fengqing Zhu\textsuperscript{1}\newline\newline 
\textsuperscript{1} School of Electrical and Computer Engineering, Purdue University, West Lafayette, Indiana, United States \newline
\textsuperscript{2}School of Public Health, Curtin University, Perth, Western Australia, Australia\newline
\textsuperscript{3}Cancer Epidemiology Program, University of Hawaii Cancer Center, Honolulu, Hawaii, United States}

\date{August 12 2020} 

\begin{document} 

\maketitle 

\thispagestyle{empty} 


\begin{abstract}
Modern deep learning techniques have enabled advances in image-based dietary assessment such as food recognition and food portion size estimation. Valuable information on the types of foods and the amount consumed are crucial for prevention of many chronic diseases. 
However, existing methods for automated image-based food analysis are neither end-to-end nor are capable of processing multiple tasks (e.g., recognition and portion estimation) together, making it difficult to apply to real life applications. In this paper, we propose an image-based food analysis framework that integrates food localization, classification and portion size estimation. Our proposed framework is end-to-end, i.e., the input can be an arbitrary food image containing multiple food items and our system can localize each single food item with its corresponding predicted food type and portion size. We also improve the single food portion estimation by consolidating localization results with a food energy distribution map obtained by conditional GAN to generate a four-channel RGB-Distribution image. Our end-to-end framework is evaluated on a real life food image dataset collected from a nutrition feeding study.
\end{abstract}

\section{Introduction}
\label{sec:intro}
Dietary assessment refers to the process of determining what someone eats and how much energy is consumed during the course of a day, which is essential for understanding the link between diet and health. Modern deep learning techniques have achieved great success in image-based dietary assessment for food localization and classification~\cite{IBM,yanai2015food,deepfood-liu2016,foodnet-pandey2017,bolanos2016simultaneous,he2020multitask, he2020incremental, mao2020visual}, as well as food portion size estimation~\cite{aizawa_2013,fang_2015,murphy_2015,dehais2018estimation,fang2019end,icip2018,he2020multitask}. However, none of these methods can achieve food localization, classification and portion size estimation in an end-to-end fashion, which makes it challenging to integrate into a complete system for fast and streamlined process. 

Image based food localization and classification problems can be viewed as specialized tasks in computer vision. The goal of food localization is to locate each individual food region for a given image with a bounding box. Pixels within the bounding box are assumed to represent a single food, which is the input to the food classification task. Food localization serves as a pre-processing step since it is common for food images in real life to contain multiple food items.

However, accurate estimation of an object's portion size is a challenging task, particularly from a single-view food image as most 3D information has been lost when the eating scene is projected from 3D world coordinates onto 2D image coordinates. An object's portion size is defined as the numeric value that is directly related to the spatial quantity of the object in world coordinates. 
The goal of food portion size estimation is to derive the food energy from an input image since energy intake is an important indicator for diet assessment.
There are existing methods~\cite{fang2019end,icip2018} that can estimate food portion size for the entire input image by generating a food energy distribution map, however, they cannot estimate the portion size of each food item separately. This is important as an individual food item can vary greatly in the energy contribution leading to significant estimation error. In this work, we address this problem by using a four-channel RGB-Distribution image, where the individual energy distribution map is obtained by applying food localization results on the entire food energy distribution map generated using conditional GAN as described in Section~\ref{method:single food portion size estimation}.

The success of modern deep learning based methods also rely on the availability of training data. 
Currently, there is no available food image dataset that includes groundtruth bounding box information, food category and corresponding portion size value for each food item in the image. Groundtruth portion size information is difficult to obtain from crowd-based annotation on RGB images, unless these numeric values are recorded during image collection. To address this issue, we introduce an eating occasion dataset containing all the groundtruth information listed above and the food portion size is provided by registered dietitians. We will describe the collection of this dataset in Section~\ref{datasetcollection}.

The main contributions of this paper can be summarized as follows.
\begin{itemize}
  \item We propose an end-to-end framework for image-based diet assessment that integrates food localization, classification and portion size estimation
  \item We introduce a novel method for single food item portion size estimation by using a four-channel RGB-Distribution image, where the individual energy distribution map is obtained by applying food localization results on the entire food energy distribution map generated by conditional GAN 
  \item We introduce a new food eating occasion image dataset containing bounding box information, food category and portion size for evaluating the proposed end-to-end framework
\end{itemize}

\begin{figure}[htbp]
\includegraphics[width=1.\linewidth]{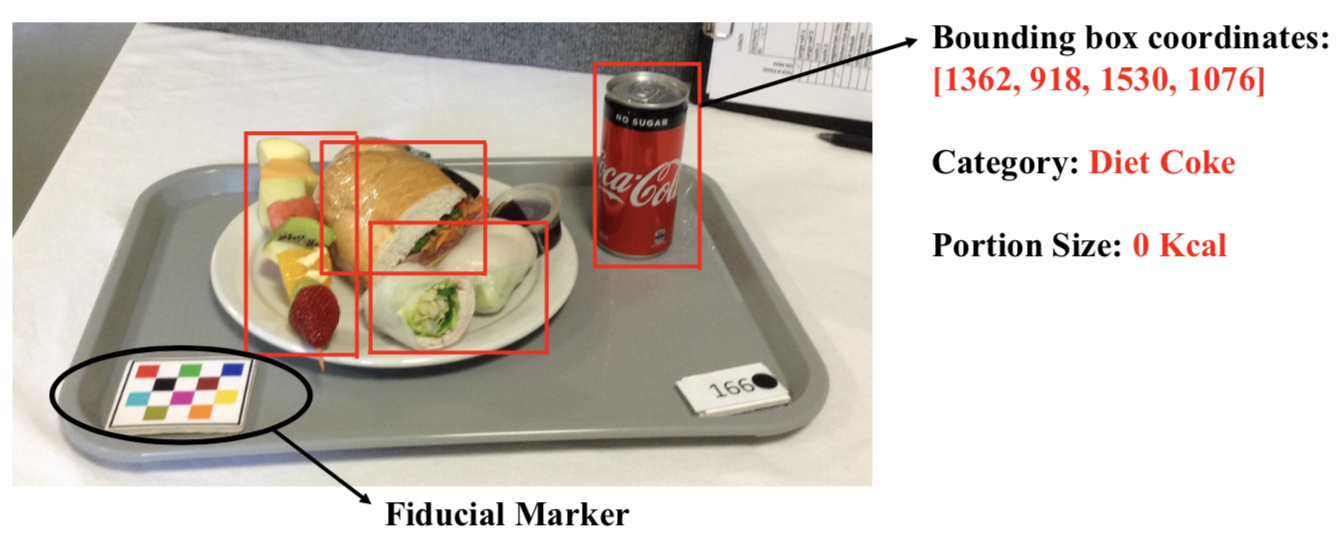}
\caption{\textbf{Example of an eating occasion image in our dataset:}
each food item is manually cropped containing the groundtruth bounding box coordinates and food category. All food and beverages were pre-weighed.  A fiducial marker is used to calibrate the color and size of the input image~\cite{FM}.}
\label{fig:dataset_example}
\end{figure}

\section{Eating Occasion Image to Food Energy Dataset}
\label{datasetcollection}
Annotated image datasets have been instrumental for driving progress in many deep learning based applications such as food detection and classification. Existing food images datasets may contain groundtruth bounding box and food label information~\cite{kawano2014automatic,Matsuda:2012ab} or just the food label~\cite{bossard14,upmc} which is not suitable for portion size estimation due to the lack of  groundtruth information. In this paper, we introduce an eating occasion image to food energy dataset containing bounding box information, food category and portion size value. Food images were collected from a nutrition study as part of an image-assisted 24-hour dietary recall (24HR) study~\cite{feedingstudy} conducted by registered dietitians. The study participants were healthy volunteers aged between 18 and 70 years old. A mobile app was used to capture images of the eating scenes for 3 meals (breakfast, lunch and dinner) over a 24-hour period. Foods are provided in buffet style where pre-weighed foods and beverages are served to the participants. Based on the known foods and their weight, food energy is calculated and used as groundtruth. The dataset contains 154 annotated eating occasion images, with a total of 915 individual food images which belong to 31 categories.  The corresponding groundtruth information includes bounding box to locate individual food, food category and portion size (in Kcal). The bounding box is given by the coordinates of input image as $[x_1, y_1, x_2, y_2]$ as shown in Figure~\ref{fig:dataset_example}.

\subsection{Data Augmentation}
We split the dataset with $15\%$ for validation $15\%$ for testing and the remaining for training. The problem with a  small dataset is that the models trained on them cannot generalize well for data from the validation and test set. Hence, these models suffer from the problem of overfitting. Data augmentation is an efficient way to address this problem, where we increase the amount of training data by rotation (90 degrees, 270 degrees) and flip (x-axis, y-axis, both). We randomly implemented the operations based on the number of training images for that category, i.e. we implemented less operations for the category which contains more images. We augment the training data while keeping the groundtruth information unchanged before and after the augmentation operations.

\begin{figure*}[htbp]

\includegraphics[width=1.\linewidth]{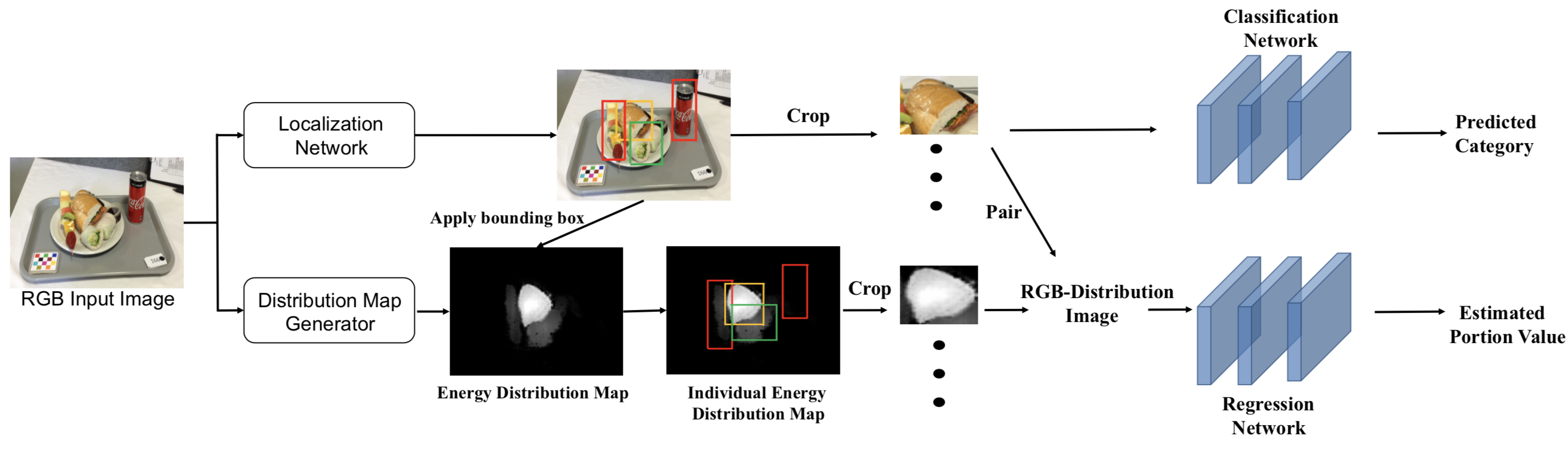}
\caption{\textbf{The overview of our proposed end-to-end framework that integrates food localization, classification and portion size estimation.} Given an input eating occasion image, the localization network locates each individual food item by generating a bounding box around the food region. Meanwhile, an energy distribution map is generated using conditional GAN. Then we directly apply a generated bounding box on an energy distribution map to get a corresponding energy distribution map for each food item. The cropped RGB food image is sent to a classification network to predict the food category. It is also used to generate the four-channel RGB-Distribution image by pairing the cropped RGB image with an individual energy distribution map, which are sent to a regression network to estimate portion size value.}
\label{Fig:overall_structure}
\end{figure*}


\section{Method}
\label{Method}
\subsection{Food Localization and Classification}
\label{method: local and class}
The goal of food localization is to locate individual food region for a given input image by providing a bounding box, where each bounding box should contain only one food item. Deep learning based methods for localization such as Faster R-CNN~\cite{ren2015faster} have shown success in many computer vision applications. It proposes potential regions that may contain the object with bounding boxes. Advanced CNN architectures such as VGG~\cite{vgg} and ResNet~\cite{resnet} can be used as the backbone structure for these methods. 

The localization network locates all individual food items within the input food image and then sends them to the classification network. We apply Convolutional Neural Networks (CNNs) to classify the food item within each bounding box, which has been widely used in image classification applications. We use cross-entropy loss $\mathcal{L}_c$ for classification task as shown below:
\begin{equation}
    \begin{centering}
    \label{eq:classification}
    \mathcal{L}_c  = \sum_{i=1}^{n}-\hat{y}^{(i)}log[f_{c}^{(i)}(\textbf{x})]
    \end{centering}
\end{equation}
where $\textbf{x}$ is the cropped food image and $\hat{y}$ is its corresponding one hot label for the food category, $f_c$ denotes the output of classification with dimension $n$.
The food localization and classification pipeline are described in Figure~\ref{Fig:overall_structure}.

\subsection{Food Portion Size Estimation}
\label{method:single food portion size estimation} 
Portion size is a property that strongly relates to the presence of an object in 3D space, so it is very difficult to accurately estimate an object's portion size by given an arbitrary 2D image. In \cite{icip2018}, a synthetic intermediate result of `energy distribution' image was proposed, where the `energy distribution' image has pixel-to-pixel correspondence and weights at different pixel locations to represent how food energy is distributed in the eating occasion. For example, pixels corresponding to steak have much higher weights than pixels of apple. \cite{Fang2019AnEI} then uses the generated distribution image to estimate food portion size by applying a regression network. On the other hand, \cite{he2020multitask} uses RGB food image only and apply feature adaptation to estimate food portion size. Our method combines the two methods and use a RGB-Distribution image to improve the estimate of the food portion size. 

\textbf{Generate energy distribution map: }We first train an energy distribution map generator by using a Generative Adversarial Networks \cite{gan} under conditional settings~\cite{pix2pix}.
We define:
\begin{equation}
G^* = \arg \min_G \max_D \mathcal{L}_{cGAN}(G, D) + \lambda \mathcal{L}_{L1}(G)
\label{eq:final_loss}
\end{equation}
where $G$ is the generator, $D$ is the discriminator,  $\mathcal{L}_{L1}(G)$ is the L1 reconstruction loss, and $\mathcal{L}_{cGAN}(G, D)$ is the conditional GAN loss as defined in \cite{pix2pix}: 
\begin{equation}
\begin{aligned}
\mathcal{L}_{cGAN}(G, D) & = \mathbb{E}_{\mathbf{x}, \mathbf{y} \sim p_{data}(\mathbf{x}, \mathbf{y})}[\log D(\mathbf{x}, \mathbf{y})] + \\
& \mathbb{E}_{\mathbf{x} \sim p_{data}(\mathbf{x}), \mathbf{z} \sim p_{z}(\mathbf{z})} [\log(1 - D(\mathbf{x}, G(\mathbf{x}, \mathbf{z}))]
\end{aligned}
\label{eq:cGANloss}
\end{equation}
where $\mathbf{x}$ is the source domain (RGB image), $\mathbf{y}$ is the target domain (energy distribution map) and $\mathbf{z}$ is random noise. The energy distribution map is a single-channel image where higher pixel value indicates higher energy distribution.

\textbf{Apply food localization bounding box: }After we generate the energy distribution map for the entire eating occasion food image, we apply the bounding box generated in Section~\ref{method: local and class} to obtain the energy distribution map for individual food item.

\textbf{Generate RGB-Distribution image: }We then combine the cropped RGB single food image with its corresponding energy distribution map to generate a RGB-Distribution image, which has four channels: R, G, B, and distribution map. The RGB-Distribution image is sent to a regression network to estimate food portion size. L1-norm loss $\mathcal{L}_r$ is used for portion size estimation:
\begin{equation}
    \label{eq:regression}
    \mathcal{L}_r  = |\hat{y} - f_{r}(x)|
\end{equation}
where $\hat{y}$ is the groundtruth portion size value and $f_r$ denotes the output of regression network with dimension 1.
The lower half of Figure~\ref{Fig:overall_structure} shows the pipeline for estimating portion size for each individual food item.

\begin{figure*}[!t]
\centering
\hspace{.5cm}
\begin{subfigure}[t]{0.31\linewidth}
    \centering
    \includegraphics[width = 4.9 cm]{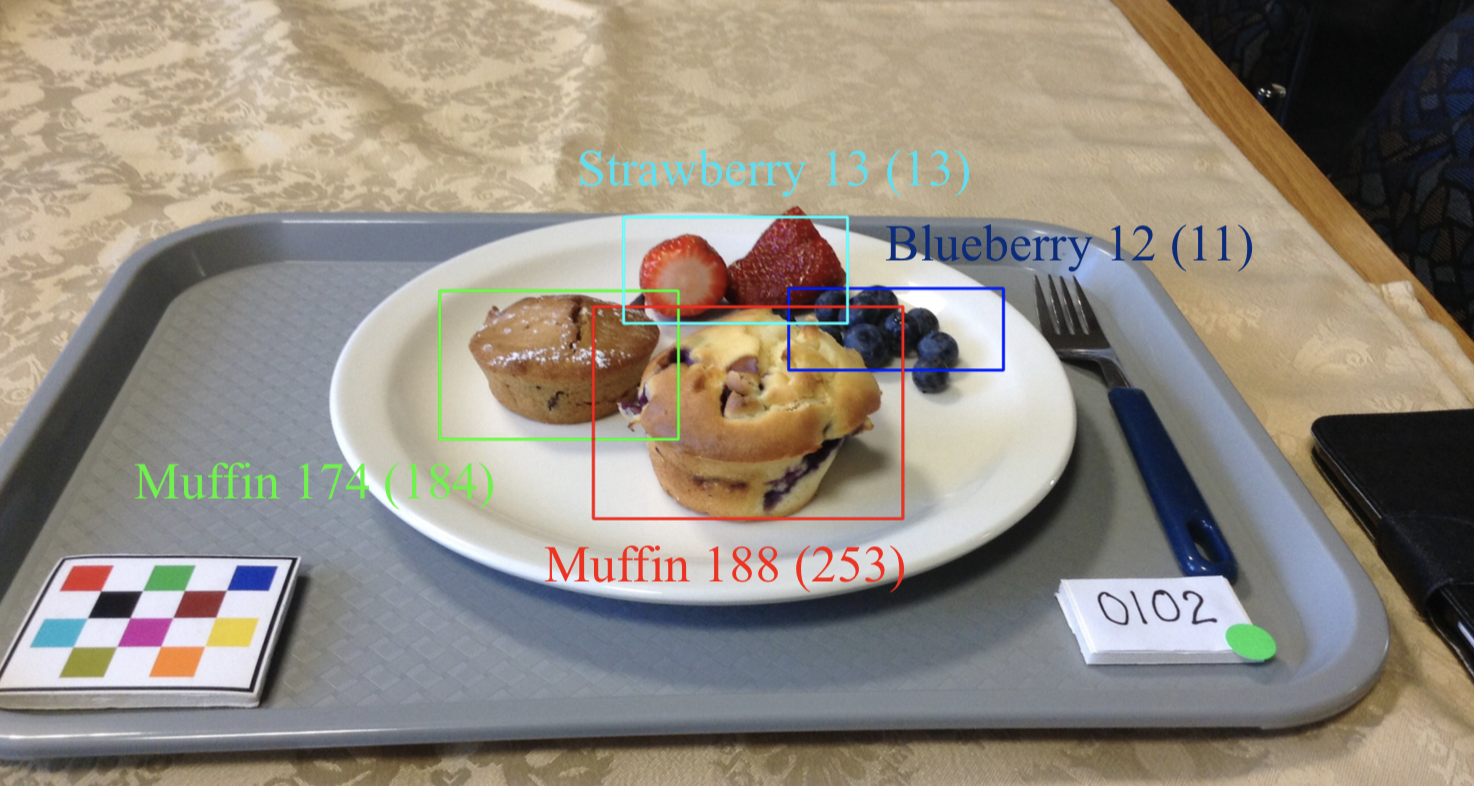}
    \caption{}\label{fig:sample-2}
\end{subfigure}
\hspace{0.1cm}
\begin{subfigure}[t]{0.31\linewidth}
    \centering
    \includegraphics[width = 4.9 cm]{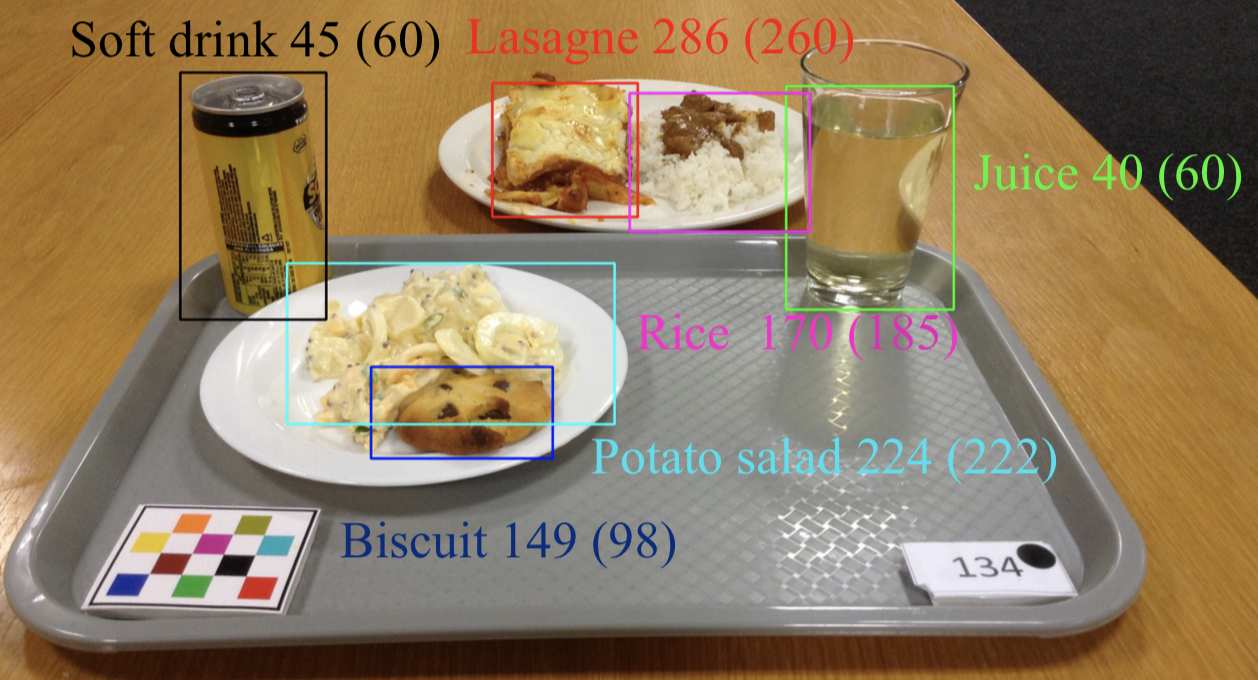}
    \caption{}\label{fig:sample-3}
\end{subfigure}
\hspace{0.1cm}
\begin{subfigure}[t]{0.31\linewidth}
    \centering
    \includegraphics[width = 4.9 cm]{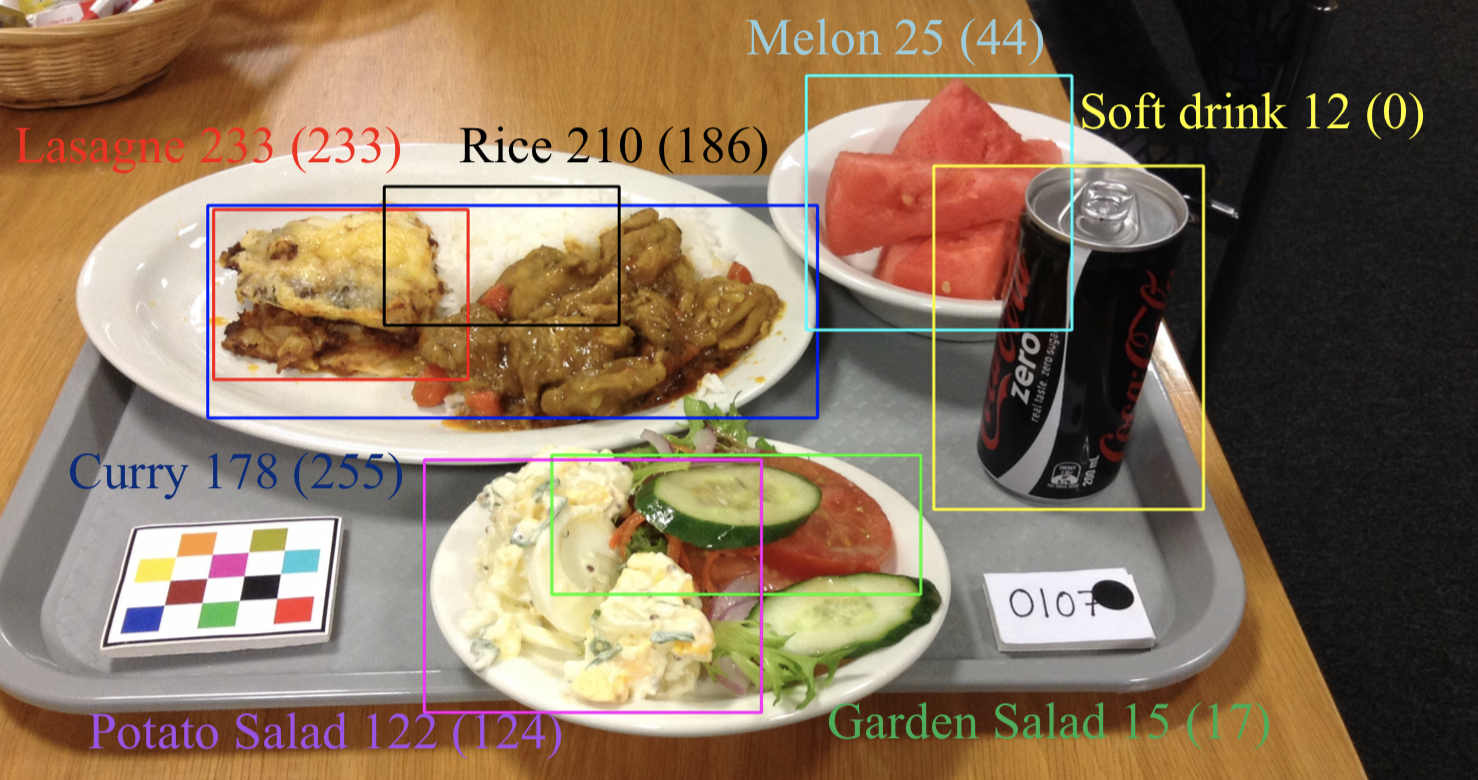}
    \caption{}\label{fig:sample-1}
\end{subfigure}
\begin{center}
\caption{\textbf{Sample results for our proposed end-to-end image analysis system.} The input to our system is a RGB food or eating scene image and the output contains the bounding box for each single food item along with the predicted category and portion size in unit of Kcal. The value inside () shows the groundtruth portion size. (Best viewed in color)}
\label{result-sample}
\end{center}
\end{figure*}

\section{Experimental Results}
In this section, we evaluate our proposed end-to-end framework using the dataset introduced in Section~\ref{datasetcollection}. 

For the localization and classification tasks, mean Average Precision (mAP) is the most common performance metrics. We firstly define several related terminologies: The intersection of union (IoU) refers to the ratio of overlapped region between predicted bounding box and groundtruth bounding box over the union of the two bounding boxes. True Positive (TP), False Positive (FP), True Negative (TN) and False Negative (FN). For example, TP means the predicted bounding box is assigned with correct food label and the corresponding IoU socre is larger than a threshold. Based on these definitions, we can calculate precision (Equation~\ref{eq:precision}) and recall (Equation~\ref{eq:recall}). 
\begin{equation} 
    \label{eq:precision}
    \text{Precision} = \frac{TP}{TP + FP}
    \end{equation}
    \begin{equation} \label{eq:recall}
    \text{Recall} = \frac{TP}{TP + FN}
\end{equation}

Average Precision (AP) for each category is the average precision value for recall value over $0$ to $1$ for each food category, and mAP is the mean value of all APs of all categories. 

Since we use L1-norm loss as shown in Equation~\ref{eq:regression} to train the regression network, we use the Mean Absolute Error (MAE) to evaluate portion size estimation, defined as 
\begin{equation}
\text{MAE} = \frac{1}{N}\sum^{N}_{i=1} |w_i - \bar{w}_i|
\label{eq:mae}
\end{equation}
where $w_i$ is the estimated portion size of the $i$-th image, $\bar{w}_i$ is the groundtruth portion size of the $i$-th image and $N$ is the number of testing images.

\subsection{Implementation Detail}
Our implementation is based on Pytorch~\cite{pytorch}. ResNet-50 is used as the backbone of Faster R-CNN. For regression network, a standard 18-layer ResNet is applied. The ResNet implementation follows the setting suggested in~\cite{resnet}. 

\subsection{Results for localization and classification}
The mAP results for food localization and classification tasks on our proposed dataset under different thresholds are shown in Table~\ref{table:loc and cls}. $0.5$ is commonly used and practical IoU threshold and we achieve satisfactory but $0.75$ is a challenging threshold as we set the threshold of $IoU > 0.75$. In addition, our dataset is challenging since the number of training data is insufficient although some data augmentation methods are implemented. We also calculate the the mAP by changing the IoU threshold from 0.5 to 0.95 with a step size of 0.05 as shown in last column.
\begin{table}[htbp]
\centering
\begin{tabular}{|c|c|c|}
\hline
mAP@.5 & mAP@.75 & mAP@[.5,.95] \\
\hline
0.6235 & 0.2428 & 0.2919\\
\hline
\end{tabular}
\caption{mAP results for food localization and classification on our introduced dataset. mAP@.5 and mAP@.75 indicate IoU larger than 0.5 and 0.75 respectively. mAP@[.5,.95] calculates AP for IoU from 0.5 to 0.95 with step size of 0.05.}
\label{table:loc and cls}
\end{table}

\subsection{Results for portion size estimation}

\textbf{Compare to state-of-the-art methods:} We compare our result of food portion size estimation with two state-of-the-art food portion estimation methods: \cite{Fang2019AnEI} and \cite{he2020multitask} that directly using food distribution map or single RGB image for regression respectively as described in Section~\ref{method:single food portion size estimation}. The input for our proposed method to estimate food portion size is a generated RGB-Distribution image of cropped RGB image and cropped energy distribution image using the localization network. As shown in Table~\ref{table:por}, our method outperforms the other two methods for single food item portion size estimation with smallest MAE as our proposed method takes into consideration for both the RGB and energy distribution information.

\begin{table}[htbp]
\centering
\begin{tabular}{|c|c|}
\hline
Methods & Mean Absolute Error (MAE) \\
\hline
Fang \textit{et al.}~\cite{Fang2019AnEI} & 109.94 Kcal\\
\hline
He \textit{et al.}~\cite{he2020multitask} & 107.55 Kcal\\
\hline
Our Method & \textbf{105.64} Kcal\\
\hline
\end{tabular}
\caption{MAE results for food portion size estimation on our introduced dataset. Best result is marked in bold.}
\label{table:por}
\end{table}
 
\textbf{Compare to human estimates:} We also compare our results for food portion size estimation of the entire eating occasion image that containing multiple single food items with 15 participants’ estimates from the same study. During the data collecting time, the participants are required to estimate the portion size of the meal they just consumed in a structured interview while viewing the eating occasion images. We sum up all single food portion size estimated by our proposed method, \cite{Fang2019AnEI} , \cite{he2020multitask} and human estimates respectively for each eating occasion image. We apply error percentage as metric in this part which is defined as 
\begin{equation}
    EP = \frac{\sum_{i=0}^N|w_i-\hat{w}_i|}{\sum_{i=0}^N\hat{w}_i}\times 100\%
\end{equation}
where $w_i$ is the estimated portion size and $\hat{w}_i$ is the groundtruth portion size. 

\begin{table}[htbp]
\centering
\begin{tabular}{|c|c|}
\hline
Methods & Error Percentage \\
\hline
Human Estimates & 62.14\%\\
\hline
Fang \textit{et al.}~\cite{Fang2019AnEI} & 35.06\%\\
\hline
He \textit{et al.}~\cite{he2020multitask} & 25.32\%\\
\hline
Our Method & \textbf{11.22\%}\\
\hline
\end{tabular}
\caption{Error percentage for food portion size estimation. Best result is marked in bold.}
\label{table:human}
\end{table}

\begin{figure}[htbp]
\includegraphics[width=1.\linewidth]{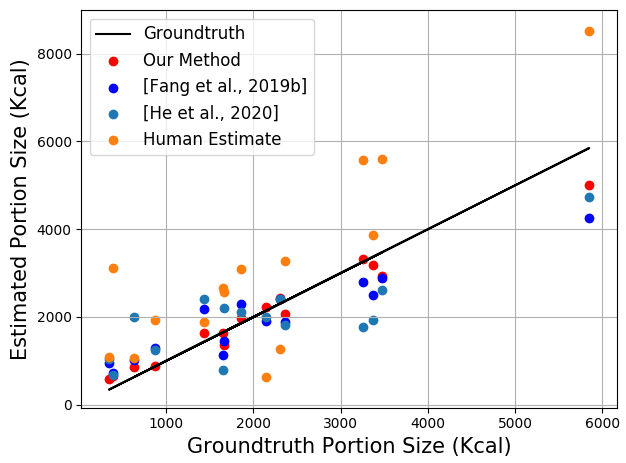}
\caption{Food portion size estimation result for each eating occasion image in test set, where the dash line indicates the groundtruth and estimated energy are the same. The dots in different color shows the results for using different methods. (Best viewed in color)}
\label{Fig:human}
\end{figure}

As shown in Table~\ref{table:human}, the error percentage (EP) of human estimates is 62.14\%, which also indicates that predicting food portion size using only information from food images is really an challenging task for majority of people. Our method gives the best result on EP for 11.22\%, which improves more than 50\% in terms of EP compared with human estimates. Figure~\ref{Fig:human} shows the results for each eating occasion image in test set. Our predicted energy (red dots) is most closest to the groundtruth energy (black line). 

\section{Conclusion}
In this paper, we propose an end-to-end image-based food analysis framework that integrates food localization, classification and portion size estimation. We introduce a novel method to estimate individual food portion size using RGB-Distribution image, where the individual energy distribution map is obtained by applying localization results on the entire energy distribution map generated by conditional GAN. Our framework is evaluated on a real life eating occasion food image dataset with groundtruth information of bounding box, food category and portion size. For localization and classification, we calculate the mAP under different thresholds and we show a satisfactory result. Our proposed method for food portion size estimation outperforms existing methods in terms of MAE as we consider both the RGB information and energy distribution information when estimating the portion size using a regression network. Our method also achieves the best improvement of error percentage from 62.14\% to 11.22\% when compared with human estimates for the entire eating occasion image, showing great potential for advancing the field of image-based dietary assessment.

\bibliographystyle{IEEEtran}
\bibliography{ref}

\begin{thebibliography}{10}
\providecommand{\url}[1]{#1}
\def\UrlFont{\rmfamily}
\providecommand{\newblock}{\relax}
\providecommand{\bibinfo}[2]{#2}
\providecommand\BIBentrySTDinterwordspacing{\spaceskip=0pt\relax}
\providecommand\BIBentryALTinterwordstretchfactor{4}
\providecommand\BIBentryALTinterwordspacing{\spaceskip=\fontdimen2\font plus
\BIBentryALTinterwordstretchfactor\fontdimen3\font minus
  \fontdimen4\font\relax}
\providecommand\BIBforeignlanguage[2]{{%
\expandafter\ifx\csname l@#1\endcsname\relax
\typeout{** WARNING: IEEEtran.bst: No hyphenation pattern has been}%
\typeout{** loaded for the language `#1'. Using the pattern for}%
\typeout{** the default language instead.}%
\else
\language=\csname l@#1\endcsname
\fi
#2}}

\bibitem{IBM}
H.~Wu, M.~Merler, R.~Uceda-Sosa, and J.~R. Smith, ``Learning to make better
  mistakes: Semantics-aware visual food recognition,'' \emph{Proceedings of the
  24th ACM international conference on Multimedia}, pp. 172--176, 2016.

\bibitem{yanai2015food}
K.~Yanai and Y.~Kawano, ``Food image recognition using deep convolutional
  network with pre-training and fine-tuning,'' \emph{Proceedings of the IEEE
  International Conference on Multimedia \& Expo Workshops}, pp. 1--6, July
  2015.

\bibitem{deepfood-liu2016}
C.~Liu, Y.~Cao, Y.~Luo, G.~Chen, V.~Vokkarane, and Y.~Ma, ``Deepfood: Deep
  learning-based food image recognition for computer-aided dietary
  assessment,'' \emph{International Conference on Smart Homes and Health
  Telematics}, pp. 37--48, 2016.

\bibitem{foodnet-pandey2017}
P.~Pandey, A.~Deepthi, B.~Mandal, and N.~B. Puhan, ``Foodnet: Recognizing foods
  using ensemble of deep networks,'' \emph{IEEE Signal Processing Letters},
  vol.~24, no.~12, pp. 1758--1762, 2017.

\bibitem{bolanos2016simultaneous}
M.~Bola{\~n}os and P.~Radeva, ``Simultaneous food localization and
  recognition,'' \emph{2016 23rd International Conference on Pattern
  Recognition}, pp. 3140--3145, 2016.

\bibitem{he2020multitask}
J.~He, Z.~Shao, J.~Wright, D.~Kerr, C.~Boushey, and F.~Zhu, ``Multi-task
  image-based dietary assessment for food recognition and portion size
  estimation,'' \emph{arXiv preprint arXiv:2004.13188}, 2020.

\bibitem{he2020incremental}
J.~He, R.~Mao, Z.~Shao, and F.~Zhu, ``Incremental learning in online
  scenario,'' \emph{Proceedings of the IEEE Conference on Computer Vision and
  Pattern Recognition}, pp. 13\,926--13\,935, 2020.

\bibitem{mao2020visual}
R.~Mao, J.~He, Z.~Shao, S.~K. Yarlagadda, and F.~Zhu, ``Visual aware hierarchy
  based food recognition,'' \emph{arXiv preprint arXiv:2012.03368}, 2020.

\bibitem{aizawa_2013}
K.~Aizawa, Y.~Maruyama, H.~Li, and C.~Morikawa, ``Food balance estimation by
  using personal dietary tendencies in a multimedia {Food Log},'' \emph{IEEE
  Transactions on Multimedia}, vol.~15, no.~8, pp. 2176 -- 2185, December 2013.

\bibitem{fang_2015}
S.~Fang, C.~Liu, F.~Zhu, E.~Delp, and C.~Boushey, ``{Single-}view food portion
  estimation based on geometric models,'' \emph{Proceedings of the IEEE
  International Symposium on Multimedia}, pp. 385--390, December 2015, {Miami,
  FL}.

\bibitem{murphy_2015}
A.~Myers, N.~Johnston, V.~Rathod, A.~Korattikara, A.~Gorban, N.~Silberman,
  S.~Guadarrama, G.~Papandreou, J.~Huang, and K.~Murphy, ``{Im2Calories:}
  towards an automated mobile vision food diary,'' \emph{Proceedings of the
  IEEE International Conference on Computer Vision}, December 2015, {Santiago,
  Chile}.

\bibitem{dehais2018estimation}
J.~Dehais, A.~Greenburg, S.~Shevchick, A.~Soni, M.~Anthimpoulos, and
  S.~Mougiakakou, ``Estimation of food volume and carbs,'' \emph{Google
  Patents}, Feb.~13 2018, uS Patent 9,892,501.

\bibitem{fang2019end}
S.~Fang, Z.~Shao, D.~A. Kerr, C.~J. Boushey, and F.~Zhu, ``An end-to-end
  image-based automatic food energy estimation technique based on learned
  energy distribution images: Protocol and methodology,'' \emph{Nutrients},
  vol.~11, no.~4, p. 877, 2019.

\bibitem{icip2018}
S.~Fang, Z.~Shao, R.~Mao, C.~Fu, E.~J. Delp, F.~Zhu, D.~A. Kerr, and C.~J.
  Boushey, ``Single-view food portion estimation: learning image-to-energy
  mappings using generative adversarial networks,'' \emph{Proceedings of the
  IEEE International Conference on Image Processing}, pp. 251--255, October
  2018, athens, Greece.

\bibitem{FM}
C.~Xu, F.~Zhu, N.~Khanna, C.~J. Boushey, and E.~J. Delp, ``Image enhancement
  and quality measures for dietary assessment using mobile devices,''
  \emph{Computational Imaging X}, vol. 8296.\hskip 1em plus 0.5em minus
  0.4em\relax International Society for Optics and Photonics, 2012, p. 82960Q.

\bibitem{kawano2014automatic}
Y.~Kawano and K.~Yanai, ``Automatic expansion of a food image dataset
  leveraging existing categories with domain adaptation,'' \emph{Proceedings of
  European Conference on Computer Vision Workshops}, pp. 3--17, September 2014,
  {Zurich, Switzerland}.

\bibitem{Matsuda:2012ab}
Y.~Matsuda, H.~Hoashi, and K.~Yanai, ``Recognition of multiple-food images by
  detecting candidate regions,'' \emph{Proceedings of IEEE International
  Conference on Multimedia and Expo}, pp. 25--30, July 2012, {Melbourne,
  Australia}.

\bibitem{bossard14}
L.~Bossard, M.~Guillaumin, and L.~V. Gool, ``Food-101 -- mining discriminative
  components with random forests,'' \emph{Proceedings of European Conference on
  Computer Vision}, vol. 8694, pp. 446--461, September 2014, {Zurich},
  Switzerland.

\bibitem{upmc}
{Xin Wang}, D.~{Kumar}, N.~{Thome}, M.~{Cord}, and F.~{Precioso}, ``Recipe
  recognition with large multimodal food dataset,'' \emph{2015 IEEE
  International Conference on Multimedia Expo Workshops (ICMEW)}, pp. 1--6,
  June 2015.

\bibitem{feedingstudy}
C.~Schipp, J.~Wright, C.~Boushy, E.~Delp, S.~Dhaliwal, and D.~Kerr, ``Can
  images improve portion size estimation of the asa24 image-assisted food
  recall: A controlled feeding study,'' \emph{Nutrition \& Dietetics; 75
  (Suppl. 1): 107}, 2018.

\bibitem{ren2015faster}
S.~Ren, K.~He, R.~Girshick, and J.~Sun, ``Faster r-cnn: Towards real-time
  object detection with region proposal networks,'' \emph{Proceedings of
  Advances in Neural Information Processing Systems}, pp. 91--99, December
  2015.

\bibitem{vgg}
K.~Simonyan and A.~Zisserman, ``Very deep convolutional networks for
  large-scale image recognition,'' \emph{arXiv preprint arXiv:1409.1556}, 2014.

\bibitem{resnet}
K.~He, X.~Zhang, S.~Ren, and J.~Sun, ``Deep residual learning for image
  recognition,'' \emph{Proceedisng of the IEEE Conference on Computer Vision
  and Pattern Recognition}, pp. 770--778, June 2016, {Las Vegas, NV}.

\bibitem{Fang2019AnEI}
S.~Fang, Z.~Shao, D.~A. Kerr, C.~J. Boushey, and F.~Zhu, ``An end-to-end
  image-based automatic food energy estimation technique based on learned
  energy distribution images: Protocol and methodology,'' \emph{Nutrients},
  vol.~11, 2019.

\bibitem{gan}
I.~Goodfellow, J.~Pouget-Abadie, M.~Mirza, B.~Xu, D.~Warde-Farley, S.~Ozair,
  A.~Courville, and Y.~Bengio, ``Generative adversarial nets,'' \emph{Advances
  in Neural Information Processing Systems 27}, pp. 2672--2680, December 2014,
  {Montreal, Canada}.

\bibitem{pix2pix}
P.~Isola, J.~Y. Zhu, T.~Zhou, and A.~A. Efros, ``Image-to-image translation
  with conditional adversarial networks,'' \emph{Proceedings of the IEEE
  Conference on Computer Vision and Pattern Recognition}, pp. 5967--5976, July
  2017, {Honolulu, HI}.

\bibitem{pytorch}
A.~Paszke, S.~Gross, S.~Chintala, G.~Chanan, E.~Yang, Z.~DeVito, Z.~Lin,
  A.~Desmaison, L.~Antiga, and A.~Lerer, ``Automatic differentiation in
  {PyTorch},'' \emph{Proceedings of the Advances Neural Information Processing
  Systems Workshop}, 2017.

\end{thebibliography}






\end{document}